\documentclass[conference]{IEEEtran}
\IEEEoverridecommandlockouts
\usepackage{cite}
\usepackage{multirow}
\usepackage{amsmath,amssymb,amsfonts}
\usepackage{algorithmic}
\usepackage{graphicx}
\usepackage{textcomp}
\usepackage{xcolor}
\usepackage{booktabs}
\usepackage{amssymb}
\usepackage{subfigure}
\def\BibTeX{{\rm B\kern-.05em{\sc i\kern-.025em b}\kern-.08em
    T\kern-.1667em\lower.7ex\hbox{E}\kern-.125emX}}
\begin{document}
\title{SFSegNet: Parse Freehand Sketches using Deep Fully Convolutional Networks}

\author{
\IEEEauthorblockN{Junkun Jiang$^1$$^3$, Ruomei Wang$^1$$^3$, Shujin Lin$^2$$ ^3$\IEEEauthorrefmark{1}\IEEEauthorrefmark{2}, Fei Wang$^1$$^3$}

\IEEEauthorblockA{$^1$School of Data and Computer Science, Sun Yat-Sen University, Guangzhou, China}
\IEEEauthorblockA{$^2$School of Communication and Design, Sun Yat-Sen University, Guangzhou, China}
\IEEEauthorblockA{$^3$National Engineering Research Center of Digital Life, Sun Yat-Sen University, Guangzhou, China}
}
\maketitle

\begin{abstract}
Parsing sketches via semantic segmentation is attractive but challenging, because (i) free-hand drawings are abstract with large variances in depicting objects due to different drawing styles and skills; (ii) distorting lines drawn on the touchpad make sketches more difficult to be recognized; 
(iii) the high-performance image segmentation via deep learning technologies needs enormous annotated sketch datasets during the training stage.

In this paper, we propose a Sketch-target deep FCN Segmentation Network(\emph{SFSegNet}) for automatic free-hand sketch segmentation, labeling each sketch in a single object with multiple parts. SFSegNet has an end-to-end network process between the input sketches and the segmentation results, composed of 2 parts: (i) a modified deep Fully Convolutional Network(FCN) using a reweighting strategy to ignore background pixels and classify which part each pixel belongs to; (ii) affine transform encoders that attempt to canonicalize the shaking strokes.
We train our network with the dataset that consists of 10,000 annotated sketches, to find an extensively applicable model to segment stokes semantically in one ground truth. Extensive experiments are carried out and segmentation results show that our method outperforms other state-of-the-art networks.
\end{abstract}


\section{Introduction}
Sketching is a ubiquitous communication way. It consists of fingertip-drawing strokes and gestures in visual images, expressing ideas across cultures and language barriers. With the wide application of touch screen technology, there's a valuable place for sketching, including image retrieval(Sketch-based Image Retrieval\cite{eitz2011sketch}\cite{qian2016enhancing}, fine-grained retrieval\cite{song2017deep}), 3D modeling\cite{cordier2016sketch} and shape retrieval\cite{wang2016data}\cite{li2017multi}. Above-mentioned technologies focus on categories labeling, computing similarity between the input sketch and existing classifications
. In this paper, we want to parse the sketch based on correspondence between strokes and different parts of the object, and furthermore understand meanings of parts and poses in sketches.

High-level ambiguity in sketches makes segmentation become a hard task. Unlike photos which have rich color, detailed objects and high contrast between foreground and background, sketches consist of sparse lines and massive blank space usually in black and white. Besides, different people have different drawing styles and depict lines from objects in their own ways, causing certain sketches to present various appearances. Some use subtle shading by multiple lines to make objects seem stereo. Some apply fewer strokes, which leads sketches to obtain ``open'' boundary parts. Shakes of the nib and distorts of the stroke during drawing also make obvious differences in drawing detail. The resulting distortions during drawing pose a challenge to implement segmentation methods.

To handle the above issues, we propose a Sketch-target deep FCN Segmentation Network(SFSegNet) for free-hand human sketch segmentation in semantically components labeling. In this architecture, we adopt two major constructions shown in Fig.\ref{fig:network_architecture}:

\begin{enumerate}
\item A sketch-targeted deep Fully Convolutional Network by fine-tuning(Section \ref{section:3-1}). Although it is still essentially a FCN\cite{long2015fully}, there are a number of crucial differences with the proposed model. First, we adopt the state-of-the-art classification network ResNet34\cite{he2016deep} for the segmentation task. According to the sparsity of lines in sketches, we use the reweighting way to avoid the part-blank(foreground-background) class imbalance. 
\item Affine transform encoders during max-pooling procedures(Section \ref{section:3-2}), similar with Spatial Transformer Networks(STN)\cite{jaderberg2015spatial}. We directly apply this encoder to the deep hierarchy output from convolutional layers to disentangle minor distortions of lines drawn by users.
\end{enumerate}

As for the dataset, we introduce the instance-level sketch segmentation dataset\cite{wang2018multi} extended from Huang's benchmark\cite{huang2014data}, consisting 10,000 annotated sketches, collected by both experts and non-experts depicting objects in ten categories after observing photos or simply imagining. To evaluate the performance of the proposed network architecture, we compare it with state-of-the-art image segmentation method FCN\cite{long2015fully}, LinkNet-34\cite{shvets2018automatic}, U-Net\cite{ronneberger2015u} and sketch segmentation method Huang's\cite{huang2014data}, CRFs\cite{schneider2016example} on both the introduced dataset and Huang's benchmark.

In Section 2, related works on image segmentation and recent approaches to sketch segmentation are reviewed. Section 3 introduces the proposed network SFSegNet's architecture design and explains the effect of in-network module combinations. In Section 4, we describe the experimental framework including our dataset and also demonstrate the results that we achieved. Finally, Section 5 concludes the paper.

\section{Related work}
In this section we discuss the prior work related to segmentation approaches for photos, scenes, and sketches. Both of them assign per-pixel predictions of object categories for the given image.

\subsection{Image Segmentation}
Recent state-of-the-art methods for semantic segmentation are based on the rapid development of Convolutional Neural Network(CNN), typically based on the Fully Convolutional Network(FCN) framework\cite{long2015fully}. FCN can transform a classification CNN, e.g. AlexNet\cite{krizhevsky2012imagenet}, VGG\cite{simonyan2014very} or GoogLeNet\cite{szegedy2015going} into a pixel-wise predictor with multiscale upsampling to tackle the semantic segmentation task.

For solving the problem of resolution loss associated with downsampling, Dilated convolution strategy is proposed\cite{chen2014semantic}\cite{yu2015multi}. This strategy handles multiscale convolution result to produce dense predictions from pretrained networks but it is lack of using global scene category clues. Inspired by Dilated convolution strategy, PSPNet\cite{zhao2017pyramid} adopts Spatial Pyramid Pooling that pools features in multiscale and concatenates them after convolution layers. Deeplab\cite{chen2018deeplab} presents an Atrous Spatial Pyramid Pooling that adopts large rate dilated convolutions. These approaches embed difficult scenery context features and reduce the model complexity by the hole algorithm.

Furthermore, in the network architecture, U-Net\cite{ronneberger2015u} is proposed to propagate context information with a large number of feature channels in upsampling part. U-Net uses skip connections to combine low-level feature maps with higher-level feature maps, which enables precise pixel-level localization.

Attempting to use in the real-time application, LinkNet\cite{chaurasia2017linknet} uses light encoders to gain fast segmentation capability. This method novelly links each encoder with decoder and bypasses the input of each encoder layer to the output of its corresponding decoder, aiming at recovering lost spatial information that is caused by downsampling processes. Shvets et al.\cite{shvets2018automatic} demonstrate the improvement of LinkNet, named as LinkNet-34, by using encoders based on a ResNet-type architecture in pre-trained weight to gain high-efficiency performance. In order to get precise segmentation boundaries, researchers have also tried to cascade their neural network with post-processing steps, like the application of the Conditional Random Field(CRF)\cite{chen2014semantic}\cite{schneider2016example}\cite{chen2018deeplab}.

\subsection{Sketch Segmentation and Labeling}
Several approaches for sketch segmentation have been suggested in the last decade, which can roughly be classified into stroke-based methods and object-based ones.

Stroke-based methods\cite{sezgin2001sketch}\cite{kim2006curvature}\cite{pu2009automated} focus on each stroke and each partition sketch by classifying which basic geometric components the strokes belong to, such as straight lines, circles, and arcs. Sezgin et al.\cite{sezgin2001sketch} extract stroke basic information about drawing direction and speed, assuming that extreme speed combined with high curvature typically corresponding to segmentation points. Kim et al.\cite{kim2006curvature} use curvature as an important criterion during the segmentation procedure. Their proposed approach was intended primarily for closed curves. Pu et al.\cite{pu2009automated} use Radial Basis Functions(RBFs) to leverage the direction and curvature of strokes.

Instead of targeting at the stroke level, Sun et al.\cite{sun2012free} first consider the sketch segmentation problem at the object level. Their solution is based on both the low-level perception and high-level knowledge, calculating the distance between each stroke to measure the proximity and recognizing from a large-scale clip-art database. There are several limitations of their method including heavily depend on the drawing sequence and normative lines. Huang et al.\cite{huang2014data} propose a data-driven approach, using parts of 3D models to match parts of sketches in a given category and performing a global optimization. In order to fit the input sketch into a certain 3D model, their technic needs a part-labeled 3D model repository and a sketch-based shape retrieval to estimate the viewpoint and the category of the input sketch. Similarly, classification before segmentation, Schneider et al.\cite{schneider2016example} adopt Fisher Vectors(FVs)\cite{sanchez2013image} for sketch classification and segment them at points with a high curvature. In consideration of the relations between segments, they fit the segmentation results into a CRF model to encode these relations. Different from their method, our model processes images of sketches without the need of any classification and drawing sequence, which makes data gathering convenient. Recently, Wu et al.\cite{wu2018sketchsegnet} present a Recurrent Neural Networks(RNN)-based model named \emph{SketchSegNet} to translate sequence of strokes into their semantic part labels. By adopting Sketch-RNN\cite{ha2017neural}, they generate a 57K annotated sketch dataset from a subset of \emph{QuickDraw} built by Google. The subset consists of 7 classes and about 60 sketches in each. However, from 420 sources drawn by the human to 57,000 sketches generated by machines, it's hard to keep the variance with the data augmentation method. Still, in their dataset, each class has its own ground truth, similar to Huang's\cite{huang2014data}. Our goal is to find an extensively applicable model to segment stokes semantically in one ground truth with low training costs.

\begin{figure*}[htbp]
	\centering
	\includegraphics[width=0.9\linewidth]{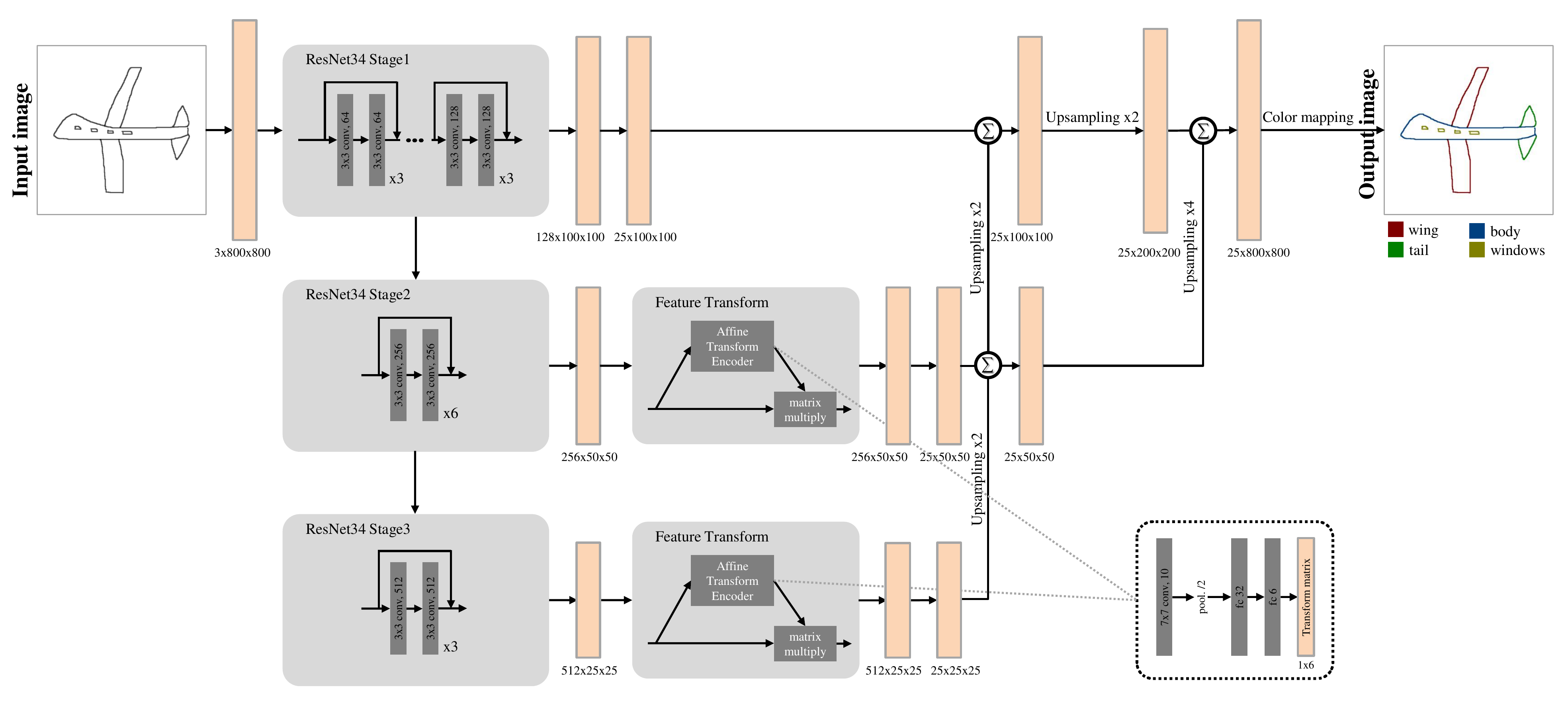}
	\caption{Illustration of the input, output and network architecture for SFSegNet. The input sketch has been amended after preprocessing.}
	\label{fig:network_architecture}
\end{figure*}

\section{Methodology}
In this section, we will first introduce the proposed network architecture of SFSegNet. The network's pipeline is shown as Fig. \ref{fig:network_architecture}, which consists of the multiscale convolutional, pooling, upsampling architecture(Section \ref{section:3-1}), and affine transform encoder(Section \ref{section:3-2}) with the reweighting strategy(Section \ref{section:fine_tune}).
Because of the characteristics of the input freehand sketch, such as sparsity, non-ordering and duplication in strokes, we cannot directly apply raw sketches to SFSegNet. Considering this, an algorithm for sketch preprocessing is described in Section \ref{section:preprocessing}.

\subsection{Architecture of SFSegNet}\label{section:3-1}
Inspired by FCN's\cite{long2015fully} multiscale learning strategy, we separate a deep CNN named as ResNet34 into 3 stages and link them for weight sharing. ResNet34 comes from the Residual Net(ResNet) family\cite{he2016deep} that consists of 34 sequential layers and has the state-of-the-art image classification performance. Each stage combines layers of features hierarchy from coarse, high level to fine, low level and gathers the necessary information. After several upsampling steps, the final output of the network is a probability map in 3-dimensional size $C\times W\times H$, indicating each probability of which part the pixel belongs to. $C$ is the number of predefined parts in segmentation or the classification number for pixels in other words. $W$ and $H$ are the shape of the input sketch, namely width and height.

In our network, we first decapitate ResNet34 by discarding the final average pooling layer and divide it into 3 stages.
We append a 2-dimension convolution layer with $C$ channels to each stage, to predict scores for each of the sketch part classes(including white background). The previous output is followed by a deconvolution layer to bilinearly upsample the coarse result to pixel-dense prediction. During training, we set the input sketch shape to $800\times800$ in RGB color format. So, a sketch in $3\times W\times H$ will lead to features under 3 different resolutions, that is, in stage1 the shape of features is $128\times(W/8)\times(H/8)$, in stage2 the shape is $256\times(W/16)\times(H/16)$ and in stage3 the shape is $128\times(W/32)\times(H/32)$. For more information, please see Table \ref{table:1}. Next, we fuse these stage results to gain more precise dense prediction. We append a 2x upsampling layer to the stage3 output and sum both the predictions computed by stage2 and stage3, notated as $s23$. Also, we append the same upsampling strategy to stage2 and combine the result and the stage1 output, notated as $s12$. We continue in this fashion by applying a 4x upsampling to the sum of predictions fused from $s12$ and $s23$. Finally, we transform the dense prediction to the sketch segmentation result.

\begin{table}[hb]
\centering
\caption{Architecture for SFSegNet. Residual blocks are building in brackets, with the numbers of blocks stacked. Upsampling is performed after each stage.}

\begin{tabular}{c|c|c}
\toprule[1pt]
stage name & output size                     &      layers             \\ \hline
\multirow{4}{*}{Stage1}           & \multirow{4}{*}{$100\times100$} & 7x7,64,stride 2 conv        \\ \cline{3-3}
           &                                 & 3x3 max pool, stride 2 \\ \cline{3-3}
           &                                 & $\begin{bmatrix}3\times3,64 \\ 3\times3,64 \end{bmatrix}\times3$ conv                      \\
           &                                 & $\begin{bmatrix}3\times3,128 \\ 3\times3,128 \end{bmatrix}\times4$ conv                      \\ \hline
Stage2     & $50\times50$                    & $\begin{bmatrix}3\times3,256 \\ 3\times3,256 \end{bmatrix}\times6$ conv                      \\ \hline
Stage3     & $25\times25$                    & $\begin{bmatrix}3\times3,512 \\ 3\times3,512 \end{bmatrix}\times3$ conv                      \\ \bottomrule[1pt]
\end{tabular}
\label{table:1}
\end{table}

\subsection{Affine Transform Encoder}\label{section:3-2}
\begin{figure}[h]
	\centering
	\includegraphics[width=0.7\linewidth]{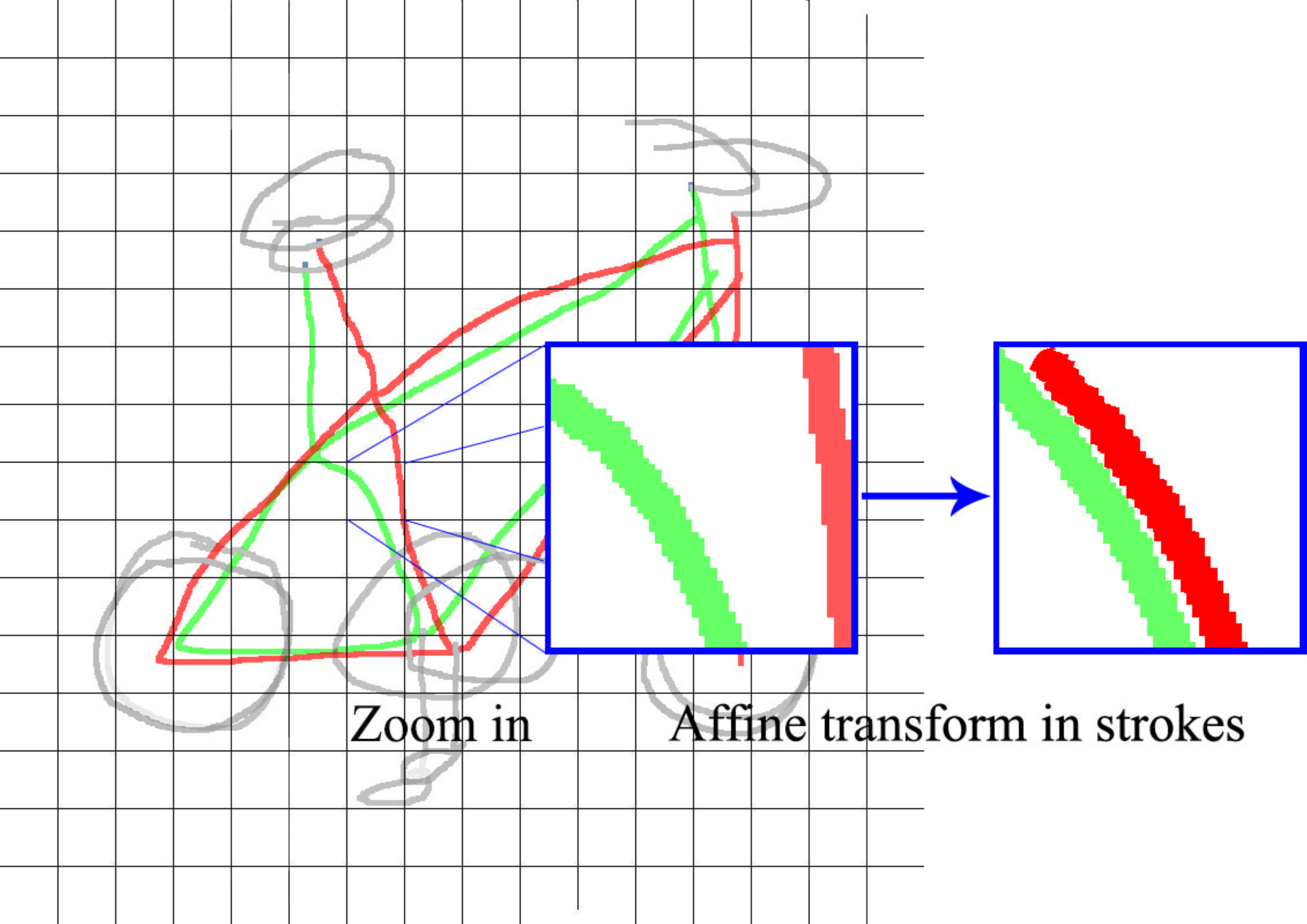}
	\caption{Sketches in one topic are drawn twice by the same volunteer. Strokes and components from two layers are hard to overlap. Blocks in the figure simulate the receptive field during convolution. Transforming positions of strokes in blocks can make a sketch's structure consistent.} 
	\label{fig:affine_transform}
\end{figure}

It's known that reproducing the same sketch during drawing is difficult. There is often a slight difference in the description of the same object due to the jitter of the strokes. This difference will have a negative effect on sketch segmentation. At the stroke level, diverse trends of strokes bring various local feature representations, and at the part level, rotation of components increases global feature differences. As shown in Fig. \ref{fig:affine_transform}, the topic of these sketches is ``bicycle'' with 6 parts. Focus on the part label ``body'', strokes in the small receptive field during convolution can be implemented with affine transform and gain spatial invariance to get better segmentation results. Moreover, with high resolutions sketches, receptive fields mostly contain only one part of the same category stroke. That means one receptive field corresponds to one stroke, which remains more structural information.

According to this thought, we employ an affine transform encoder to generate a transformation matrix to align the output of the feature maps extracted in resolutions from the lower level to the global level. The applied encoder is a mini STN\cite{jaderberg2015spatial} and enables the network to have the ability to correct hand drawn deviation. Our affine transform encoder only has one convolution layer in the localization network, different from STN which has two convolution layers but affine transformation still works towards the sampled output feature map.

\subsection{Fine-tune}\label{section:fine_tune}
Three stages extracted from ResNet34 are pre-trained on ImageNet, before fine-tuning on the sketch dataset. We use the cross-entropy function as a loss for deep model training. Given $x$ as a discrete probability distribution and $class$ as the correct class of the input, the cross-entropy function is defined as:
\begin{equation}\label{eq:1}
\text{loss}(x, class) = -x[class] + \log\left(\sum_j^C \exp(x[j])\right)
\end{equation}
where $C$ is the number of classes. In terms of the characteristics of sketches, there are several simple curves and almost blank space in a sketch. The area of strokes occupied less than 1\% according to the statistics on the dataset. In the paper, we treat the blank area, namely the background of the sketch, as one of the components to be segmented. About 99\% pixels in the same color(R:255, G:255, B:255) will be classified as ``background'' and the rest of them will be classified as about 3 to 4 categories. The unbalanced data makes the segmentation model more likely to classify all pixels as ``background'', so the segmentation result is almost in white. We have a strong reason to reweight the ``background'' class before training. The reweighting loss can be described as:
\begin{equation}\label{eq:2}
\begin{split}
\text{loss}(x, class) = weight[class](-x[class] \\ + \log\left(\sum_j^C \exp(x[j])\right))
\end{split}
\end{equation}
During training, we set the weight of ``background'' to 0 and other classes to 1, ignoring blank pixels in the loss computation. Next section will show how good results we achieve using the reweighting strategy.

\subsection{Preprocessing}\label{section:preprocessing}
To arrange the sketch as a normalized input form for training, we centralize and recolor the raw data. We first use a bounding box to enclose the sketch and resize it randomly(from $600\times600$ to $700\times700$ pixels). Resized sketches are padding to $800\times800$ pixels in the center. To avoid the impact of the interpolation algorithm during scaling, we erode the strokes into 1 pixel and recolor each pixel to correct labels.

\section{Experiments}
We evaluate our network in two datasets. Results in different datasets show the proposed technology is perfectly competent in sketch-target segmentation and detailed implementations are described as follow.

\subsection{Datasets}
\begin{figure}[htbp]
	\centering
	\includegraphics[width=0.9\linewidth]{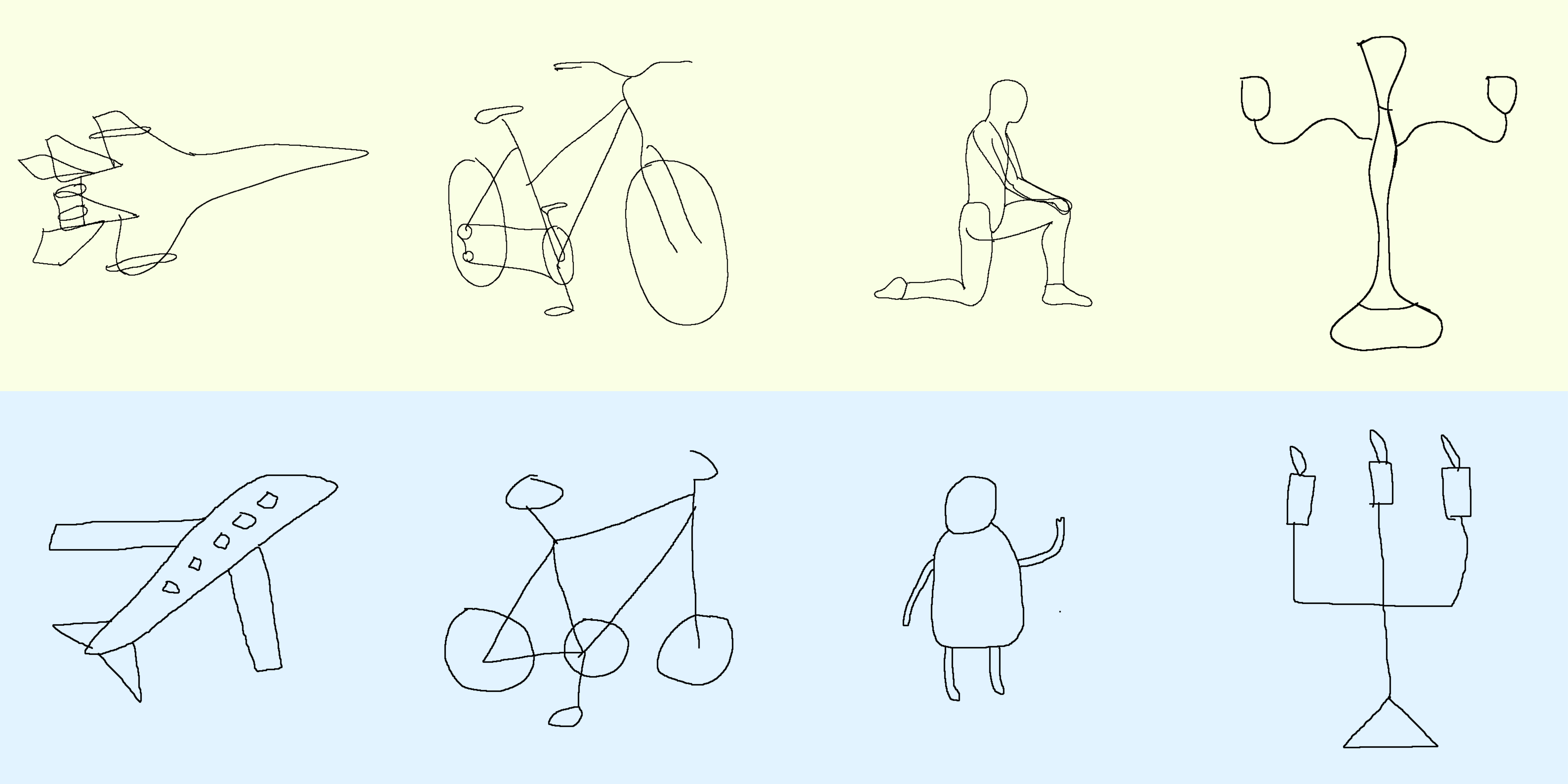}
	\caption{The first row includes several sketches from Huang's dataset, which are closely similar with 3D meshes, while sketches are freehand. The second row includes examples from our dataset, which consists of more natural strokes, such that provides more noise.}
	\label{fig:dataset_huang}
\end{figure}

We first introduce the component-labeled sketch dataset built by Huang et al.\cite{huang2014data}, which contains 10 classes and 300 sketches drawn by 3 users(i.e., for each class, 10 from each user). Their examples are created by a quick glance at a natural image, thus are much more realistic than the usual sketches which are more likely to be imaginary(see Fig. \ref{fig:dataset_huang}). Huang's dataset focuses on whether each sketch has as many labeled parts as possible, but each category has an independent set of ground truths. Scarce annotated datasets for training and single ground truth for each sketch, the above factors make Huang's dataset unsuitable for deep learning.

\begin{figure*}[htbp]
	\centering
	\includegraphics[width=1\linewidth]{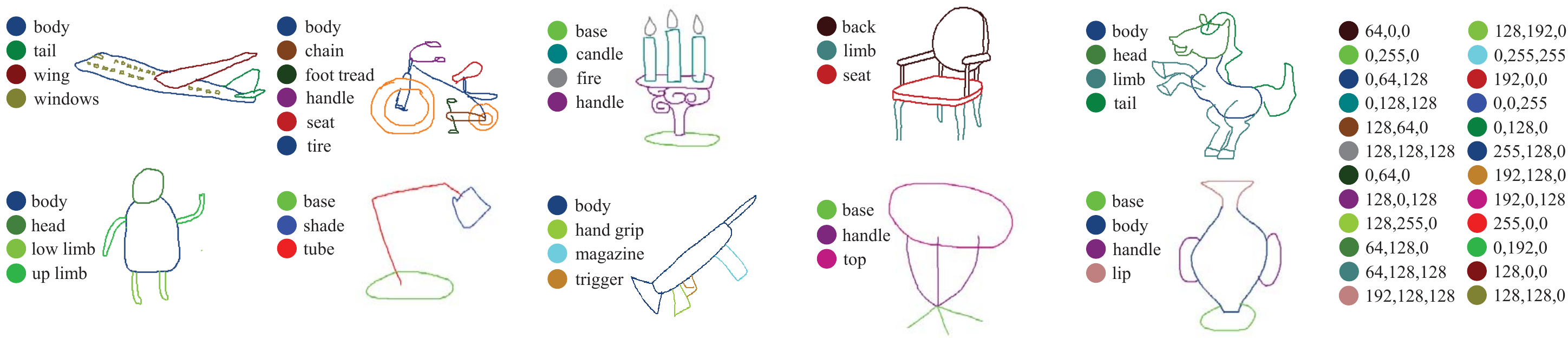}
	\caption{Examples of labeled sketches in one ground truth with the maximum number of components. The first row shows sketches drawn by experts from the Airplane, Bicycle, Candelabra, Chair, and Fourleg categories. The second row shows sketches drawn by non-experts from the Human, Lamp, Rifle, Table, and Vase categories. We tag every component with three parameters on the RGB color model, such as the airplane can be constructed by body(R:0, G:64, B:128), wing(R:128, G:0, B:0), tail(R:0, G:128, B:0), and windows(R:128, G:128, B:0).}
	\label{fig:dataset_ours}
\end{figure*}

\begin{figure*}[htbp]
	\centering
	\includegraphics[width=1\linewidth]{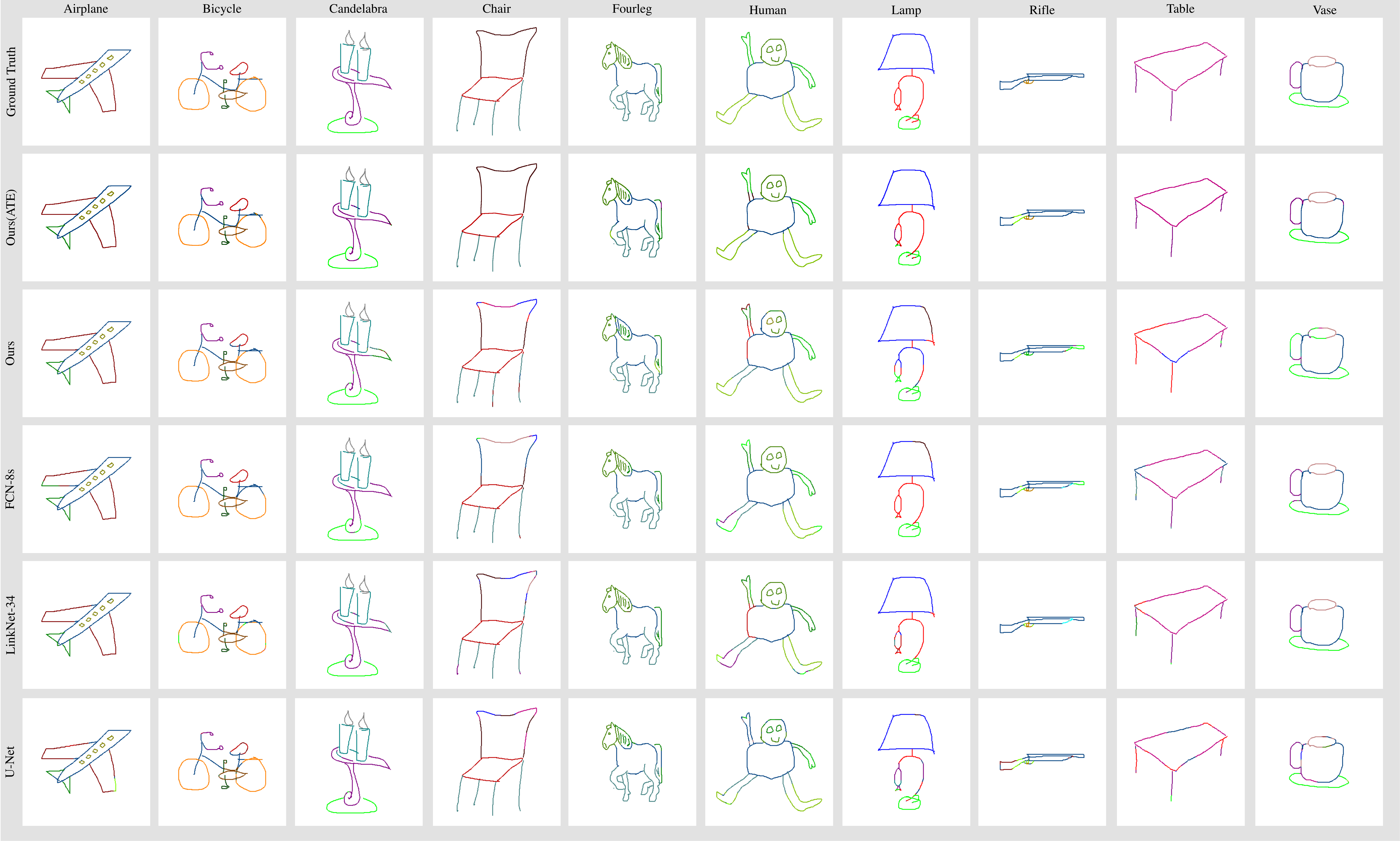}
	\caption{SFSegNet produces a state-of-the-art performance on our dataset, compared with FCN-8s, LinkNet-34, and U-Net. According to its outperformance compared with FCN-16s and FCN-32s, FCN-8s is selected for presentation. The first column shows the ground truth for our dataset. The second column shows the output of our highest performing net with affine transform encoders. The third column shows the output of our net without affine transform encoders.}
	\label{fig:dataset_ours}
\end{figure*}

Followed by Huang et al., we build a large-scale dataset\cite{wang2018multi} that consists 10,000 sketches and 25 components(include background) in one ground truth for each sketch. We taste 10 familiar classes for their easy imaginativeness: Airplane(6 components), Bicycle(5 components), Candelabra(4 components), Chair(3 components) Fourleg(4 components), Human(4 components), Lamp(3 components), Rifle(4 components), Table(3 components) and Vase(4 components). Labeled examples and components' tags on RGB space are shown in Fig. \ref{fig:dataset_ours}. For each class, there are 1,000 sketches drawn by 10 volunteers; half of them are experienced artists and the rest are not. We ask both of them to draw sketches on the digital tablet in 1 minute. The content of sketches is immediately thought up when volunteers receive a topic, for closing to much more natural representations. Though a ceiling number of components has been set, volunteers can decide how many components in sketches.

\subsection{Implementation Details}
Our model is implemented with Pytorch on a PC with a single NVIDIA 1080TI, an i5-7400 3GHz CPU and 16GB RAM. We divide our dataset into 2 subsets, 75\% for training and 25\% for testing. We utilize randomly initialized decoder weights and encoder weights initialized with ResNet34, pre-trained on ImageNet. The initial learning rate is set to 0.001, and the mini-batch size is set to 5. During training, we use stochastic gradient descent with the momentum of 0.9 and a polynomial weight decay policy. For baseline models in deep learning including FCN\cite{long2015fully}, LinkNet-34\cite{shvets2018automatic} and U-Net\cite{ronneberger2015u}, we adopt their default training parameters. Also, we optimize baselines with reweighting strategy described in Section \ref{section:fine_tune}. All models are trained within 50 iterations.

\subsection{Evaluation}
Different from image segmentation, pixels in the sketch have been pre-classified into two categories: strokes and background. It's inappropriate to use image segmentation's evaluation method such as IoU(Intersection over Union) or AP(Average Precision). To evaluate the segmentation performance for sketches, we adopt two accuracy metrics followed by Huang et al\cite{huang2014data}: 1) Pixel-based accuracy (\textbf{P-metric}), the number of pixels with correct labels divided by the total number. 2) Component-based accuracy (\textbf{C-metric}), the ratio of the number of a component with correct labels to the total number. A component is correctly labeled if the number of its correct pixels is up to 75\%.

\subsection{Results and Discussion}
\textbf{Experiments on Huang's Dataset.} We train our network on our dataset and test it on Huang's dataset. Notice that Huang's dataset has ten ground truths for each category, differing from our dataset's configuration(one ground truth for all categories). We remove and combine components to apply the same settings. It's unfair to evaluate the network on C-metric due to the inconsistent number of components. We evaluate on P-metric merely. Besides, some components are annotated by mistake. We relabel them to ensure all sketches are fine-labeled correctly. Also, the same preprocessing has been implemented before testing. The above contents will be explained in the appendix.

\begin{table}[htbp]
\centering
\vspace{-3mm}
\caption{Accuracy(\%) on Huang's dataset, using the P-metric.}
\begin{tabular}{l|c|c||c}
\toprule[1pt]
\multicolumn{1}{l|}{}           & \multicolumn{1}{c|}{Huang}         & \multicolumn{1}{c||}{CRF}           & SFSegNet     \\ \hline
\multicolumn{1}{l|}{Airplane}   & \multicolumn{1}{c|}{\textbf{74.0}} & \multicolumn{1}{c||}{55.1}          & 65.5          \\ \hline
\multicolumn{1}{l|}{Bicycle}    & \multicolumn{1}{c|}{72.6}          & \multicolumn{1}{c||}{79.7}          & \textbf{83.4} \\ \hline
\multicolumn{1}{l|}{Candelabra} & \multicolumn{1}{c|}{59.0}          & \multicolumn{1}{c||}{\textbf{72.0}} & 64.9          \\ \hline
\multicolumn{1}{l|}{Chair}      & \multicolumn{1}{c|}{52.6}          & \multicolumn{1}{c||}{\textbf{66.5}} & 63.0          \\ \hline
\multicolumn{1}{l|}{Fourleg}    & \multicolumn{1}{c|}{77.9}          & \multicolumn{1}{c||}{\textbf{81.5}} & 79.4          \\ \hline
\multicolumn{1}{l|}{Human}      & \multicolumn{1}{c|}{62.5}          & \multicolumn{1}{c||}{69.7}          & \textbf{77.0} \\ \hline
\multicolumn{1}{l|}{Lamp}       & \multicolumn{1}{c|}{82.5}          & \multicolumn{1}{c||}{82.9}          & \textbf{94.3} \\ \hline
\multicolumn{1}{l|}{Rifle}      & \multicolumn{1}{c|}{66.9}          & \multicolumn{1}{c||}{67.8}          & \textbf{80.4} \\ \hline
\multicolumn{1}{l|}{Table}      & \multicolumn{1}{c|}{67.9}          & \multicolumn{1}{c||}{\textbf{74.5}} & 61.4          \\ \hline
\multicolumn{1}{l|}{Vase}       & \multicolumn{1}{c|}{63.2}          & \multicolumn{1}{c||}{\textbf{83.3}} & 73.1          \\ \hline
\multicolumn{1}{l|}{Average}    & \multicolumn{1}{c|}{67.9}          & \multicolumn{1}{c||}{73.2}          & \textbf{74.2} \\ \bottomrule[1pt]
\multicolumn{4}{l}{\textit{Note: Best results are in boldface.}}
\end{tabular}
\label{table:2}
\vspace{-2mm}
\end{table}

Table \ref{table:2} shows that our method outperforms the Huang's method\cite{huang2014data} and has similar performance but much less test time-consuming(1 to 2 sketches per second) compared with the CRF model\cite{schneider2016example}. However, towards certain categories, SFSegNet is about 20\% less than CRF. The most likely cause of unsatisfactory accuracy results is, CRF has categorization information for each sketch. Without the prior knowledge, it's a tough work to classify strokes by only relying on grouping information. For example, some instances in class ``Candelabra'' have more than one candles which are far apart from each other. The local feature representation brings less correlation between them. Shown in Fig. \ref{fig:dataset_huang}, the lamp-like handle makes the network easily recognize it as a ``lamp'' object, though we are pretty sure that a candelabra couldn't have a lamp inside.

\textbf{Experiments on Our Dataset.} We report the comparative performance of our network SFSegNet and other methods including FCN\cite{long2015fully}, LinkNet-34\cite{shvets2018automatic}, U-Net\cite{ronneberger2015u} as baselines, due to their successfully application in semantic segmentation. We also discuss the effect of the affine transform encoder. To avoid the background label biasing normal labels during training, all models use the reweighting loss. Segmentation results are shown in Fig. \ref{fig:dataset_ours}.

\begin{table}[htbp]
\centering
\caption{Accuracy(\%) on our dataset, using the P-metric.}
\vspace{-1mm}
\setlength{\tabcolsep}{0.8mm}{
\begin{tabular}{lcccccc}
\toprule[1pt]
\multicolumn{1}{l|}{}           & \multicolumn{1}{c|}{FCN-8s} & \multicolumn{1}{c|}{FCN-16s} & \multicolumn{1}{c|}{FCN-32s} & \multicolumn{1}{c|}{LinkNet-34}    & \multicolumn{1}{c||}{U-Net} & SFSegNet     \\ \hline
\multicolumn{1}{l|}{Airplane}   & \multicolumn{1}{c|}{91.1}   & \multicolumn{1}{c|}{91.2}    & \multicolumn{1}{c|}{90.5}    & \multicolumn{1}{c|}{92.3}          & \multicolumn{1}{c||}{80.9}  & \textbf{93.3} \\ \hline
\multicolumn{1}{l|}{Bicycle}    & \multicolumn{1}{c|}{92.8}   & \multicolumn{1}{c|}{91.9}    & \multicolumn{1}{c|}{89.8}    & \multicolumn{1}{c|}{\textbf{94.0}} & \multicolumn{1}{c||}{86.9}  & 93.3          \\ \hline
\multicolumn{1}{l|}{Candelabra} & \multicolumn{1}{c|}{92.4}   & \multicolumn{1}{c|}{91.9}    & \multicolumn{1}{c|}{91.2}    & \multicolumn{1}{c|}{92.7}          & \multicolumn{1}{c||}{89.6}  & \textbf{94.3} \\ \hline
\multicolumn{1}{l|}{Chair}      & \multicolumn{1}{c|}{86.9}   & \multicolumn{1}{c|}{85.9}    & \multicolumn{1}{c|}{85.7}    & \multicolumn{1}{c|}{87.7}          & \multicolumn{1}{c||}{79.4}  & \textbf{90.3} \\ \hline
\multicolumn{1}{l|}{Fourleg}    & \multicolumn{1}{c|}{86.6}   & \multicolumn{1}{c|}{86.2}    & \multicolumn{1}{c|}{84.9}    & \multicolumn{1}{c|}{88.1}          & \multicolumn{1}{c||}{83.8}  & \textbf{89.2} \\ \hline
\multicolumn{1}{l|}{Human}      & \multicolumn{1}{c|}{83.7}   & \multicolumn{1}{c|}{82.9}    & \multicolumn{1}{c|}{81.8}    & \multicolumn{1}{c|}{\textbf{85.9}} & \multicolumn{1}{c||}{77.0}  & 85.1          \\ \hline
\multicolumn{1}{l|}{Lamp}       & \multicolumn{1}{c|}{86.6}   & \multicolumn{1}{c|}{87.2}    & \multicolumn{1}{c|}{86.2}    & \multicolumn{1}{c|}{90.1}          & \multicolumn{1}{c||}{87.5}  & \textbf{91.4} \\ \hline
\multicolumn{1}{l|}{Rifle}      & \multicolumn{1}{c|}{88.6}   & \multicolumn{1}{c|}{87.7}    & \multicolumn{1}{c|}{87.2}    & \multicolumn{1}{c|}{90.3}          & \multicolumn{1}{c||}{84.7}  & \textbf{91.1} \\ \hline
\multicolumn{1}{l|}{Table}      & \multicolumn{1}{c|}{85.7}   & \multicolumn{1}{c|}{84.2}    & \multicolumn{1}{c|}{85.1}    & \multicolumn{1}{c|}{86.7}          & \multicolumn{1}{c||}{81.0}  & \textbf{90.7} \\ \hline
\multicolumn{1}{l|}{Vase}       & \multicolumn{1}{c|}{91.2}   & \multicolumn{1}{c|}{90.9}    & \multicolumn{1}{c|}{90.3}    & \multicolumn{1}{c|}{90.3}          & \multicolumn{1}{c||}{87.9}  & \textbf{93.5} \\ \hline
\multicolumn{1}{l|}{Average}    & \multicolumn{1}{c|}{88.6}   & \multicolumn{1}{c|}{88.0}    & \multicolumn{1}{c|}{87.3}    & \multicolumn{1}{c|}{89.8}          & \multicolumn{1}{c||}{83.9}  & \textbf{91.2} \\ \bottomrule[1pt]
\multicolumn{7}{l}{\textit{Note: Best results are in boldface.}}
\end{tabular}}
\label{table:ours_P}
\vspace{-3mm}
\end{table}

\begin{table}[htbp]
\centering
\caption{Accuracy(\%) on our dataset, using the C-metric.}
\vspace{-1mm}
\setlength{\tabcolsep}{0.8mm}{
\begin{tabular}{lcccccc}
\toprule[1pt]
\multicolumn{1}{l|}{}           & \multicolumn{1}{c|}{FCN-8s} & \multicolumn{1}{c|}{FCN-16s} & \multicolumn{1}{c|}{FCN-32s} & \multicolumn{1}{c|}{LinkNet-34} & \multicolumn{1}{c||}{U-Net}         & SFSegNet      \\ \hline
\multicolumn{1}{l|}{Airplane}   & \multicolumn{1}{c|}{85.0}   & \multicolumn{1}{c|}{85.3}    & \multicolumn{1}{c|}{83.8}    & \multicolumn{1}{c|}{78.9}       & \multicolumn{1}{c||}{86.4}          & \textbf{87.3} \\ \hline
\multicolumn{1}{l|}{Bicycle}    & \multicolumn{1}{c|}{85.2}   & \multicolumn{1}{c|}{83.8}    & \multicolumn{1}{c|}{80.3}    & \multicolumn{1}{c|}{74.6}       & \multicolumn{1}{c||}{\textbf{85.9}} & 85.6          \\ \hline
\multicolumn{1}{l|}{Candelabra} & \multicolumn{1}{c|}{94.1}   & \multicolumn{1}{c|}{93.1}    & \multicolumn{1}{c|}{92.8}    & \multicolumn{1}{c|}{83.2}       & \multicolumn{1}{c||}{94.1}          & \textbf{95.3} \\ \hline
\multicolumn{1}{l|}{Chair}      & \multicolumn{1}{c|}{86.9}   & \multicolumn{1}{c|}{85.4}    & \multicolumn{1}{c|}{86.1}    & \multicolumn{1}{c|}{83.4}       & \multicolumn{1}{c||}{88.4}          & \textbf{89.6} \\ \hline
\multicolumn{1}{l|}{Fourleg}    & \multicolumn{1}{c|}{84.1}   & \multicolumn{1}{c|}{83.3}    & \multicolumn{1}{c|}{81.7}    & \multicolumn{1}{c|}{71.5}       & \multicolumn{1}{c||}{86.3}          & \textbf{87.5} \\ \hline
\multicolumn{1}{l|}{Human}      & \multicolumn{1}{c|}{81.6}   & \multicolumn{1}{c|}{81.1}    & \multicolumn{1}{c|}{79.3}    & \multicolumn{1}{c|}{74.3}       & \multicolumn{1}{c||}{\textbf{85.7}} & 82.8          \\ \hline
\multicolumn{1}{l|}{Lamp}       & \multicolumn{1}{c|}{87.1}   & \multicolumn{1}{c|}{88.0}    & \multicolumn{1}{c|}{87.5}    & \multicolumn{1}{c|}{83.3}       & \multicolumn{1}{c||}{91.2}          & \textbf{92.1} \\ \hline
\multicolumn{1}{l|}{Rifle}      & \multicolumn{1}{c|}{82.0}   & \multicolumn{1}{c|}{80.3}    & \multicolumn{1}{c|}{80.0}    & \multicolumn{1}{c|}{72.4}       & \multicolumn{1}{c||}{\textbf{83.4}} & 83.2          \\ \hline
\multicolumn{1}{l|}{Table}      & \multicolumn{1}{c|}{82.2}   & \multicolumn{1}{c|}{79.6}    & \multicolumn{1}{c|}{81.5}    & \multicolumn{1}{c|}{72.2}       & \multicolumn{1}{c||}{84.6}          & \textbf{87.5} \\ \hline
\multicolumn{1}{l|}{Vase}       & \multicolumn{1}{c|}{92.2}   & \multicolumn{1}{c|}{92.4}    & \multicolumn{1}{c|}{91.9}    & \multicolumn{1}{c|}{84.7}       & \multicolumn{1}{c||}{91.2}          & \textbf{94.9} \\ \hline
\multicolumn{1}{l|}{Average}    & \multicolumn{1}{c|}{86.0}   & \multicolumn{1}{c|}{85.2}    & \multicolumn{1}{c|}{84.5}    & \multicolumn{1}{c|}{77.9}       & \multicolumn{1}{c||}{87.7}          & \textbf{88.6} \\ \bottomrule[1pt]
\multicolumn{7}{l}{\textit{Note: Best results are in boldface.}}
\end{tabular}}
\label{table:ours_C}
\vspace{-1mm}
\end{table}

\begin{figure*}[htbp]
\begin{center}
\subfigure[Reweighting strategy activated.]{ \label{fig:reweighting_a} \includegraphics[width = 0.4\linewidth]{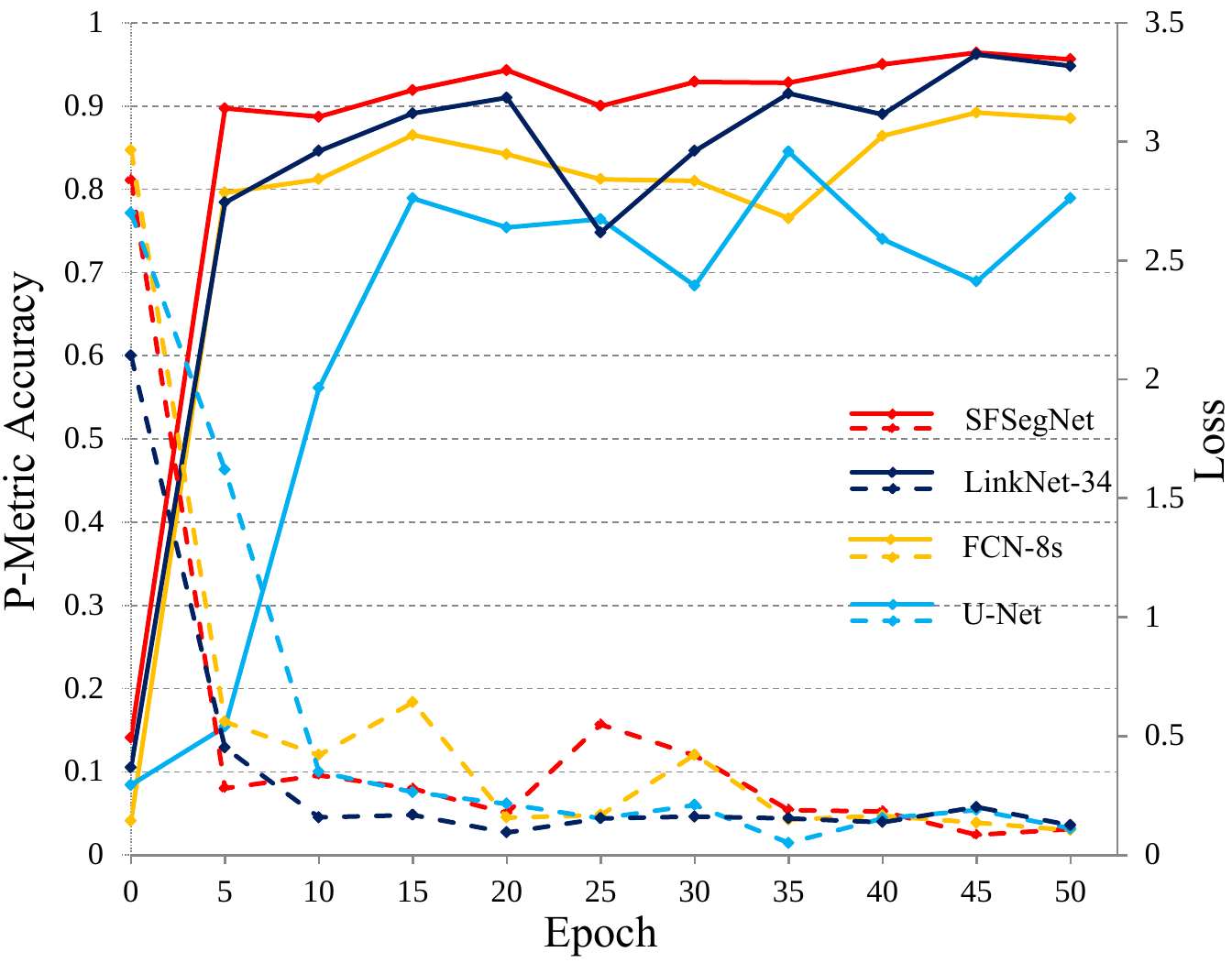}}
\hspace{23mm}
\subfigure[Reweighting strategy deactivated.]{ \label{fig:reweighting_b} \includegraphics[width = 0.4\linewidth]{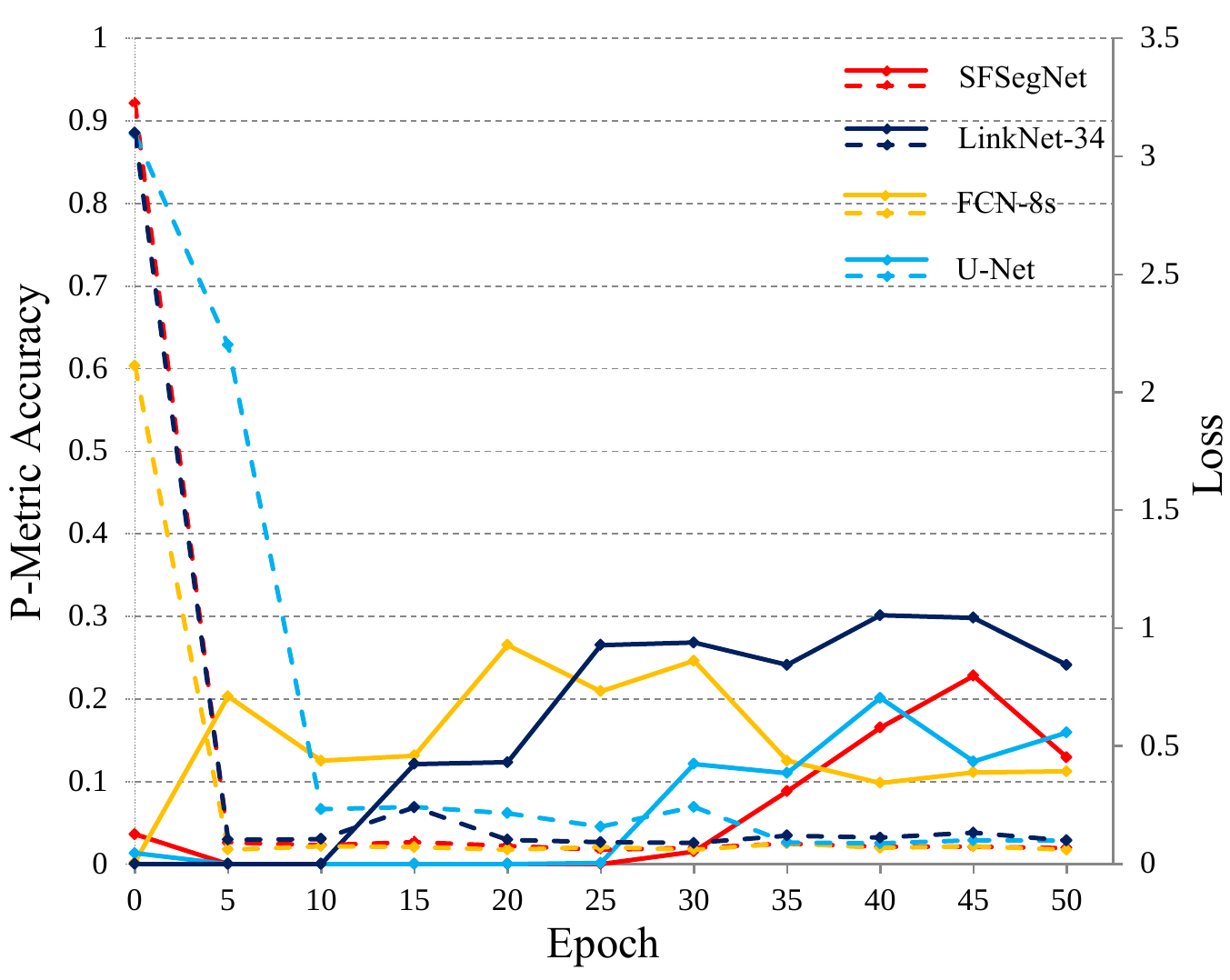}}
\end{center}
\vspace{-3mm}
\caption{Comparison of segmentation results}
\label{fig:reweighting}
\vspace{-3mm}
\end{figure*}

\subsubsection{Reweighting strategy}Quantitative results including the segmentation accuracy based on P-metric and loss during training are shown in Fig.\ref{fig:reweighting}. We can observe from Fig.\ref{fig:reweighting_b} that, when the reweighting strategy has been deactivated, the loss of each network is decreasing, however, the accuracy is increasing slightly and eventually stops at a low level. Suffering heavily from heavy class imbalance problem, the model predicts pixels in one component which produces meaningless results.
And Fig.\ref{fig:reweighting_a} indicates that the reweighing strategy can solve the problem and speed up fitting. Each of them gains high accuracy within about 10 epoch.

\subsubsection{Baselines}
The quantitative results of the proposed network and the competitors are presented in Table \ref{table:ours_P} and \ref{table:ours_C}. The average labeling accuracy for each class is presented. We can observe that our model performs the best in each metric. Using P-metric, the average accuracy is 2.9\% higher than FCN-8s, 3.5\% higher than FCN-16s, 3.5\% higher than FCN-32s, 1.5\% higher than the LinkNet and 8.0\% higher than the U-net. Using P-metric, the average accuracy is 2.9\% higher than FCN-8s, 3.8\% higher than FCN-16s, 4.6\% higher than FCN-32s, 12.1\% higher than the LinkNet and 1.0\% higher than the U-net.

\begin{table}[htbp]
\centering
\caption{Accuracy(\%) on our dataset, compared with the application of the affine transform encoder(ATE).}
\begin{tabular}{lcccc}
\toprule[1pt]
\multicolumn{1}{l|}{}           & \multicolumn{2}{c|}{P-metric}                                  & \multicolumn{2}{c}{C-metric}              \\ \hline
\multicolumn{1}{l|}{ATE}        & \multicolumn{1}{c|}{}     & \multicolumn{1}{c|}{$\checkmark$}              & \multicolumn{1}{c|}{}     &    {$\checkmark$}              \\ \hline
\multicolumn{1}{l|}{Airplane}   & \multicolumn{1}{c|}{92.4} & \multicolumn{1}{c|}{\textbf{93.3}} & \multicolumn{1}{c|}{86.7} & \textbf{87.3} \\ \hline
\multicolumn{1}{l|}{Bicycle}    & \multicolumn{1}{c|}{93.1} & \multicolumn{1}{c|}{\textbf{93.3}} & \multicolumn{1}{c|}{84.1} & \textbf{85.6} \\ \hline
\multicolumn{1}{l|}{Candelabra} & \multicolumn{1}{c|}{92.8} & \multicolumn{1}{c|}{\textbf{94.3}} & \multicolumn{1}{c|}{94.1} & \textbf{95.3} \\ \hline
\multicolumn{1}{l|}{Chair}      & \multicolumn{1}{c|}{87.3} & \multicolumn{1}{c|}{\textbf{90.3}} & \multicolumn{1}{c|}{86.2} & \textbf{89.6} \\ \hline
\multicolumn{1}{l|}{Fourleg}    & \multicolumn{1}{c|}{86.3} & \multicolumn{1}{c|}{\textbf{89.2}} & \multicolumn{1}{c|}{83.0} & \textbf{87.5} \\ \hline
\multicolumn{1}{l|}{Human}      & \multicolumn{1}{c|}{83.5} & \multicolumn{1}{c|}{\textbf{85.1}} & \multicolumn{1}{c|}{82.0} & \textbf{82.8} \\ \hline
\multicolumn{1}{l|}{Lamp}       & \multicolumn{1}{c|}{88.5} & \multicolumn{1}{c|}{\textbf{91.4}} & \multicolumn{1}{c|}{90.3} & \textbf{92.1} \\ \hline
\multicolumn{1}{l|}{Rifle}      & \multicolumn{1}{c|}{88.6} & \multicolumn{1}{c|}{\textbf{91.1}} & \multicolumn{1}{c|}{80.6} & \textbf{83.2} \\ \hline
\multicolumn{1}{l|}{Table}      & \multicolumn{1}{c|}{87.5} & \multicolumn{1}{c|}{\textbf{90.7}} & \multicolumn{1}{c|}{85.7} & \textbf{87.5} \\ \hline
\multicolumn{1}{l|}{Vase}       & \multicolumn{1}{c|}{91.5} & \multicolumn{1}{c|}{\textbf{93.5}} & \multicolumn{1}{c|}{92.3} & \textbf{94.9} \\ \hline
\multicolumn{1}{l|}{Average}    & \multicolumn{1}{c|}{89.2} & \multicolumn{1}{c|}{\textbf{91.2}} & \multicolumn{1}{c|}{86.5} & \textbf{88.6} \\ \bottomrule[1pt]
\multicolumn{5}{l}{\textit{Note: Best results are in boldface.}}
\end{tabular}
\label{table:ATE}
\end{table}

\subsubsection{Affine transform encoder}
We remove all affine transform encoders from our SFSegNet to discuss their effect. Table \ref{table:ATE} shows the comparison results. Three stages of SFSegNet learn strokes structural features from hierarchy layers and obtain good results in segmentation, even better than baselines. We note that some components composed of straight strokes but labeled more than two categories. We believe there is a strong possibility that shaking strokes bring noise to convolutional features and make the components' probability map predict more than one part. With spatial invariance during convolution, the strokes' features are canonicalized and our model can achieve a better segmentation result.

\section{Conclusion}
In this paper, a sketch-targeted deep network named SFSegNet is proposed. We observe the class imbalance through blank labels and component labels. By using a reweighting strategy during training, the background pixels are ignored and the overall structure information in part-wise is well preserved. To prevent the disturbance caused by shaking strokes, we apply the affine transform encoder to gain spatial invariance during convolution to get more robust features. Essentially, it learns from the structural information among the drawing strokes. Thus the fully convolutional decoder is able to get the better segmentation results. Experimental results validated the effectiveness of our proposed method.

\appendix
\renewcommand{\appendixname}{Appendix~\Alph{section}}
In this paper, components from Huang's dataset\cite{huang2014data} are removed or combined to make sketches consistent with our dataset. Configurations are shown in Table \ref{table:config}.

\begin{table}[]
\centering
\caption{Dataset configuration}
\setlength{\tabcolsep}{0.8mm}{
\begin{tabular}{l|llll|l|l|l|l}
\toprule[1pt]
                            & \multicolumn{4}{c|}{Huang}                                                                                      & \multicolumn{4}{c}{Huang(with our config)}                                                     \\ \cline{2-9}
                            & \multicolumn{1}{l|}{Components} & \multicolumn{1}{c|}{R}   & \multicolumn{1}{c|}{G}   & \multicolumn{1}{c|}{B} & Components      & \multicolumn{1}{c|}{R}   & \multicolumn{1}{c|}{G}   & \multicolumn{1}{c}{B}   \\ \hline
\multirow{6}{*}{Airplane}   & \multicolumn{1}{l|}{body}       & \multicolumn{1}{c|}{255} & \multicolumn{1}{c|}{0}   & \multicolumn{1}{c|}{0} & body            & \multicolumn{1}{c|}{0}   & \multicolumn{1}{c|}{64}  & \multicolumn{1}{c}{128} \\ \cline{2-9}
                            & \multicolumn{1}{l|}{wing}       & \multicolumn{1}{l|}{0}   & \multicolumn{1}{l|}{255} & 0                      & wing            & 128                      & 0                        & 0                       \\ \cline{2-9}
                            & \multicolumn{1}{l|}{horistab}   & \multicolumn{1}{l|}{0}   & \multicolumn{1}{l|}{0}   & 255                    & tail            & 0                        & 128                      & 0                       \\ \cline{2-9}
                            & \multicolumn{1}{l|}{vertstab}   & \multicolumn{1}{l|}{255} & \multicolumn{1}{l|}{255} & 0                      & tail            & 0                        & 128                      & 0                       \\ \cline{2-9}
                            & \multicolumn{1}{l|}{engine}     & \multicolumn{1}{l|}{255} & \multicolumn{1}{l|}{0}   & 255                    & \textit{ignore} &                          &                          &                         \\ \cline{2-9}
                            & \multicolumn{1}{l|}{propeller}  & \multicolumn{1}{l|}{0}   & \multicolumn{1}{l|}{255} & 255                    & \textit{ignore} &                          &                          &                         \\ \hline
\multirow{9}{*}{Bicycle}    & \multicolumn{1}{l|}{saddle}     & \multicolumn{1}{c|}{255} & \multicolumn{1}{c|}{0}   & \multicolumn{1}{c|}{0} & seat            & \multicolumn{1}{c|}{192} & \multicolumn{1}{c|}{0}   & \multicolumn{1}{c}{0}   \\ \cline{2-9}
                            & \multicolumn{1}{l|}{frontframe} & \multicolumn{1}{l|}{0}   & \multicolumn{1}{l|}{255} & 0                      & body            & 0                        & 64                       & 128                     \\ \cline{2-9}
                            & \multicolumn{1}{l|}{wheel}      & \multicolumn{1}{l|}{0}   & \multicolumn{1}{l|}{0}   & 255                    & tire            & 255                      & 128                      & 0                       \\ \cline{2-9}
                            & \multicolumn{1}{l|}{handle}     & \multicolumn{1}{l|}{255} & \multicolumn{1}{l|}{255} & 0                      & handle          & 128                      & 0                        & 128                     \\ \cline{2-9}
                            & \multicolumn{1}{l|}{pedal}      & \multicolumn{1}{l|}{255} & \multicolumn{1}{l|}{0}   & 255                    & foottread       & 0                        & 64                       & 0                       \\ \cline{2-9}
                            & \multicolumn{1}{l|}{chain}      & \multicolumn{1}{l|}{0}   & \multicolumn{1}{l|}{255} & 255                    & chain           & 128                      & 64                       & 0                       \\ \cline{2-9}
                            & \multicolumn{1}{l|}{fork}       & \multicolumn{1}{l|}{128} & \multicolumn{1}{l|}{0}   & 0                      & body            & 0                        & 64                       & 128                     \\ \cline{2-9}
                            & \multicolumn{1}{l|}{backframe}  & \multicolumn{1}{l|}{0}   & \multicolumn{1}{l|}{128} & 0                      & body            & 0                        & 64                       & 128                     \\ \cline{2-9}
                            & \multicolumn{1}{l|}{backcover}  & \multicolumn{1}{l|}{0}   & \multicolumn{1}{l|}{0}   & 128                    & body            & 0                        & 64                       & 128                     \\ \hline
\multirow{6}{*}{Candelabra} & \multicolumn{1}{l|}{base}       & \multicolumn{1}{c|}{255} & \multicolumn{1}{c|}{0}   & \multicolumn{1}{c|}{0} & base            & \multicolumn{1}{c|}{0}   & \multicolumn{1}{c|}{255} & \multicolumn{1}{c}{0}   \\ \cline{2-9}
                            & \multicolumn{1}{l|}{candle}     & \multicolumn{1}{l|}{0}   & \multicolumn{1}{l|}{255} & 0                      & candle          & 0                        & 128                      & 128                     \\ \cline{2-9}
                            & \multicolumn{1}{l|}{fire}       & \multicolumn{1}{l|}{0}   & \multicolumn{1}{l|}{0}   & 255                    & fire            & 128                      & 128                      & 128                     \\ \cline{2-9}
                            & \multicolumn{1}{l|}{handle}     & \multicolumn{1}{l|}{255} & \multicolumn{1}{l|}{255} & 0                      & handle          & 128                      & 0                        & 128                     \\ \cline{2-9}
                            & \multicolumn{1}{l|}{shaft}      & \multicolumn{1}{l|}{255} & \multicolumn{1}{l|}{0}   & 255                    & handle          & 128                      & 0                        & 128                     \\ \cline{2-9}
                            & \multicolumn{1}{l|}{arm}        & \multicolumn{1}{l|}{0}   & \multicolumn{1}{l|}{255} & 255                    & handle          & 128                      & 0                        & 128                     \\ \hline
\multirow{11}{*}{Chair}     & \multicolumn{1}{l|}{back}       & \multicolumn{1}{c|}{255} & \multicolumn{1}{c|}{0}   & \multicolumn{1}{c|}{0} & back            & \multicolumn{1}{c|}{64}  & \multicolumn{1}{c|}{0}   & \multicolumn{1}{c}{0}   \\ \cline{2-9}
                            & \multicolumn{1}{l|}{leg}        & \multicolumn{1}{l|}{0}   & \multicolumn{1}{l|}{255} & 0                      & limb            & 64                       & 128                      & 128                     \\ \cline{2-9}
                            & \multicolumn{1}{l|}{seat}       & \multicolumn{1}{l|}{0}   & \multicolumn{1}{l|}{0}   & 255                    & seat            & 192                      & 0                        & 0                       \\ \cline{2-9}
                            & \multicolumn{1}{l|}{arm}        & \multicolumn{1}{l|}{255} & \multicolumn{1}{l|}{255} & 0                      & limb            & 64                       & 128                      & 128                     \\ \cline{2-9}
                            & \multicolumn{1}{l|}{stile}      & \multicolumn{1}{l|}{255} & \multicolumn{1}{l|}{0}   & 255                    & back            & 64                       & 0                        & 0                       \\ \cline{2-9}
                            & \multicolumn{1}{l|}{gas lift}   & \multicolumn{1}{l|}{0}   & \multicolumn{1}{l|}{255} & 255                    & limb            & 64                       & 128                      & 128                     \\ \cline{2-9}
                            & \multicolumn{1}{l|}{base}       & \multicolumn{1}{l|}{128} & \multicolumn{1}{l|}{0}   & 0                      & limb            & 64                       & 128                      & 128                     \\ \cline{2-9}
                            & \multicolumn{1}{l|}{foot}       & \multicolumn{1}{l|}{0}   & \multicolumn{1}{l|}{128} & 0                      & limb            & 64                       & 128                      & 128                     \\ \cline{2-9}
                            & \multicolumn{1}{l|}{stretcher}  & \multicolumn{1}{l|}{0}   & \multicolumn{1}{l|}{0}   & 128                    & limb            & 64                       & 128                      & 128                     \\ \cline{2-9}
                            & \multicolumn{1}{l|}{spindle}    & \multicolumn{1}{l|}{128} & \multicolumn{1}{l|}{128} & 0                      & back            & 64                       & 0                        & 0                       \\ \cline{2-9}
                            & \multicolumn{1}{l|}{rail}       & \multicolumn{1}{l|}{0}   & \multicolumn{1}{l|}{128} & 128                    & back            & 64                       & 0                        & 0                       \\ \hline
\multirow{5}{*}{Fourleg}    & \multicolumn{1}{l|}{body}       & \multicolumn{1}{c|}{255} & \multicolumn{1}{c|}{0}   & \multicolumn{1}{c|}{0} & body            & \multicolumn{1}{c|}{0}   & \multicolumn{1}{c|}{64}  & \multicolumn{1}{c}{128} \\ \cline{2-9}
                            & \multicolumn{1}{l|}{ear}        & \multicolumn{1}{l|}{0}   & \multicolumn{1}{l|}{255} & 0                      & head            & 64                       & 128                      & 0                       \\ \cline{2-9}
                            & \multicolumn{1}{l|}{head}       & \multicolumn{1}{l|}{0}   & \multicolumn{1}{l|}{0}   & 255                    & head            & 64                       & 128                      & 0                       \\ \cline{2-9}
                            & \multicolumn{1}{l|}{leg}        & \multicolumn{1}{l|}{255} & \multicolumn{1}{l|}{255} & 0                      & limb            & 64                       & 128                      & 128                     \\ \cline{2-9}
                            & \multicolumn{1}{l|}{tail}       & \multicolumn{1}{l|}{255} & \multicolumn{1}{l|}{0}   & 255                    & tail            & 0                        & 128                      & 0                       \\ \hline
\multirow{6}{*}{Human}      & \multicolumn{1}{l|}{head}       & \multicolumn{1}{c|}{255} & \multicolumn{1}{c|}{0}   & \multicolumn{1}{c|}{0} & head            & \multicolumn{1}{c|}{64}  & \multicolumn{1}{c|}{128} & \multicolumn{1}{c}{0}   \\ \cline{2-9}
                            & \multicolumn{1}{l|}{body}       & \multicolumn{1}{l|}{0}   & \multicolumn{1}{l|}{255} & 0                      & body            & 0                        & 64                       & 128                     \\ \cline{2-9}
                            & \multicolumn{1}{l|}{arm}        & \multicolumn{1}{l|}{0}   & \multicolumn{1}{l|}{0}   & 255                    & uplimb          & 0                        & 192                      & 0                       \\ \cline{2-9}
                            & \multicolumn{1}{l|}{leg}        & \multicolumn{1}{l|}{255} & \multicolumn{1}{l|}{255} & 0                      & lowlimb         & 128                      & 192                      & 0                       \\ \cline{2-9}
                            & \multicolumn{1}{l|}{hand}       & \multicolumn{1}{l|}{255} & \multicolumn{1}{l|}{0}   & 255                    & uplimb          & 0                        & 192                      & 0                       \\ \cline{2-9}
                            & \multicolumn{1}{l|}{foot}       & \multicolumn{1}{l|}{0}   & \multicolumn{1}{l|}{255} & 255                    & lowlimb         & 128                      & 192                      & 0                       \\ \hline
\multirow{3}{*}{Lamp}       & \multicolumn{1}{l|}{tube}       & \multicolumn{1}{c|}{255} & \multicolumn{1}{c|}{0}   & \multicolumn{1}{c|}{0} & tube            & \multicolumn{1}{c|}{255} & \multicolumn{1}{c|}{0}   & \multicolumn{1}{c}{0}   \\ \cline{2-9}
                            & \multicolumn{1}{l|}{base}       & \multicolumn{1}{l|}{0}   & \multicolumn{1}{l|}{255} & 0                      & base            & 0                        & 255                      & 0                       \\ \cline{2-9}
                            & \multicolumn{1}{l|}{shade}      & \multicolumn{1}{l|}{0}   & \multicolumn{1}{l|}{0}   & 255                    & shade           & 0                        & 0                        & 255                     \\ \hline
\multirow{7}{*}{Rifle}      & \multicolumn{1}{l|}{barrel}     & \multicolumn{1}{c|}{255} & \multicolumn{1}{c|}{0}   & \multicolumn{1}{c|}{0} & body            & \multicolumn{1}{c|}{0}   & \multicolumn{1}{c|}{64}  & \multicolumn{1}{c}{128} \\ \cline{2-9}
                            & \multicolumn{1}{l|}{body}       & \multicolumn{1}{l|}{0}   & \multicolumn{1}{l|}{255} & 0                      & body            & 0                        & 64                       & 128                     \\ \cline{2-9}
                            & \multicolumn{1}{l|}{handgrip}   & \multicolumn{1}{l|}{0}   & \multicolumn{1}{l|}{0}   & 255                    & handgrip        & 128                      & 255                      & 0                       \\ \cline{2-9}
                            & \multicolumn{1}{l|}{magazine}   & \multicolumn{1}{l|}{255} & \multicolumn{1}{l|}{255} & 0                      & magazine        & 0                        & 255                      & 255                     \\ \cline{2-9}
                            & \multicolumn{1}{l|}{trigger}    & \multicolumn{1}{l|}{255} & \multicolumn{1}{l|}{0}   & 255                    & trigger         & 192                      & 128                      & 0                       \\ \cline{2-9}
                            & \multicolumn{1}{l|}{butt}       & \multicolumn{1}{l|}{0}   & \multicolumn{1}{l|}{255} & 255                    & body            & 0                        & 64                       & 128                     \\ \cline{2-9}
                            & \multicolumn{1}{l|}{sight}      & \multicolumn{1}{l|}{128} & \multicolumn{1}{l|}{0}   & 0                      & body            & 0                        & 64                       & 128                     \\ \hline
\multirow{7}{*}{Table}      & \multicolumn{1}{l|}{top}        & \multicolumn{1}{c|}{255} & \multicolumn{1}{c|}{0}   & \multicolumn{1}{c|}{0} & top             & \multicolumn{1}{c|}{192} & \multicolumn{1}{c|}{0}   & \multicolumn{1}{c}{128} \\ \cline{2-9}
                            & \multicolumn{1}{l|}{leg}        & \multicolumn{1}{l|}{0}   & \multicolumn{1}{l|}{255} & 0                      & handle          & 128                      & 0                        & 128                     \\ \cline{2-9}
                            & \multicolumn{1}{l|}{stretcher}  & \multicolumn{1}{l|}{0}   & \multicolumn{1}{l|}{0}   & 255                    & handle          & 128                      & 0                        & 128                     \\ \cline{2-9}
                            & \multicolumn{1}{l|}{base}       & \multicolumn{1}{l|}{255} & \multicolumn{1}{l|}{255} & 0                      & base            & 0                        & 255                      & 0                       \\ \cline{2-9}
                            & \multicolumn{1}{l|}{topsupport} & \multicolumn{1}{l|}{255} & \multicolumn{1}{l|}{0}   & 255                    & handle          & 128                      & 0                        & 128                     \\ \cline{2-9}
                            & \multicolumn{1}{l|}{legsupport} & \multicolumn{1}{l|}{0}   & \multicolumn{1}{l|}{255} & 255                    & handle          & 128                      & 0                        & 128                     \\ \cline{2-9}
                            & \multicolumn{1}{l|}{midsupport} & \multicolumn{1}{l|}{128} & \multicolumn{1}{l|}{0}   & 0                      & handle          & 128                      & 0                        & 128                     \\ \hline
\multirow{4}{*}{Vase}       & \multicolumn{1}{l|}{lip}        & \multicolumn{1}{c|}{255} & \multicolumn{1}{c|}{0}   & \multicolumn{1}{c|}{0} & lip             & \multicolumn{1}{c|}{192} & \multicolumn{1}{c|}{128} & \multicolumn{1}{c}{128} \\ \cline{2-9}
                            & \multicolumn{1}{l|}{handle}     & \multicolumn{1}{l|}{0}   & \multicolumn{1}{l|}{255} & 0                      & handle          & 128                      & 0                        & 128                     \\ \cline{2-9}
                            & \multicolumn{1}{l|}{body}       & \multicolumn{1}{l|}{0}   & \multicolumn{1}{l|}{0}   & 255                    & body            & 0                        & 64                       & 128                     \\ \cline{2-9}
                            & \multicolumn{1}{l|}{foot}       & \multicolumn{1}{l|}{255} & \multicolumn{1}{l|}{255} & 0                      & base            & 0                        & 255                      & 0                       \\ \bottomrule[1pt]
\end{tabular}}
\label{table:config}
\end{table}

\bibliographystyle{IEEEtrans}
\bibliography{SketchSegBib}

\end{document}